\documentclass[10pt,twocolumn,letterpaper]{article}

\PassOptionsToPackage{table}{xcolor} 

\usepackage[pagenumbers]{wacv} 

\usepackage{booktabs}
\usepackage{multirow}
\usepackage{array}
\usepackage{tabularx}
\usepackage{amssymb}
\usepackage{pifont}
\usepackage{enumitem}
\setlist{nosep,leftmargin=*}

\newcommand{\cmark}{\ding{51}}

\newcommand{\pmark}{$\circ$}
\newcommand{\pre}{$^\dagger$}          
\newcommand{\para}[1]{\noindent\textbf{#1}\hspace{0.5em}}
\newcommand{\hns}{\mathrm{HNS}}

\definecolor{wacvblue}{rgb}{0.21,0.49,0.74}
\usepackage[pagebackref,breaklinks,colorlinks,allcolors=wacvblue]{hyperref}

\def\wacvPaperID{1} 
\def\confName{WACV}
\def\confYear{2027}

\title{The Gap Is in the Geometry: A Convergent Meta-Analysis of Multi-Facet\\Aerial and Remote Sensing Video Understanding}

\author{Georgios Angelidis\\
Harokopio University of Athens\\
Athens, Greece\\
{\tt\small gangelidis@hua.gr}
\and
Aggelos Psiris\\
Harokopio University of Athens\\
Athens, Greece\\
{\tt\small aggelospsiris@hua.gr}
\and
Giorgos Sfikas\\
University of West Attica\\
Athens, Greece\\
{\tt\small gsfikas@uniwa.gr}
\and
Maria Vakalopoulou\\
CentraleSupelec\\
Paris, France\\
{\tt\small maria.vakalopoulou@centralesupelec.fr}
\and
Iraklis Varlamis\\
Harokopio University of Athens\\
Athens, Greece\\
{\tt\small varlamis@hua.gr}
\and
Georgios Th. Papadopoulos\\
Harokopio University of Athens\\
Athens, Greece\\
{\tt\small G.Th.Papadopoulos@hua.gr}
}

\begin{document}
\maketitle

\begin{abstract}
Aerial and remote sensing video is increasingly evaluated like web video, by posing multiple-choice questions to multimodal large language models (MLLMs) over sampled frames and reporting accuracy.
This paper argues that this recipe measures the wrong aspect, and supports the claim with a convergent meta-analysis of a broad set of published results, organized in four evidence tiers, in which a claim becomes a position only when collected evidence agrees.
Parameter scale exhibits a median of $+3.4$ accuracy points per tenfold increase on aerial benchmarks, far below a human gap of at least 19 points, whereas acquisition conditions control perception, as 315 of 318 evaluations in 19 within-paper contrasts lose accuracy when conditions such as altitude, view, object size or time horizon become harder.
No aerial benchmark in the corpus combines video input with per-task human and random baselines, competence collapses from answering to acting in closed-loop navigation, and the same MLLMs score higher on the aerial benchmark than on ground spatial ones, but lower on studies that test cross-view geometry.
From these findings the current study derives five falsifiable principles and proposes the AVU-Card, a reporting card for aerial video benchmarks.
\end{abstract}

\section{Introduction}
\label{sec:intro}

Video from drones, aircraft and satellites is now routinely analyzed with the same multimodal large language models (MLLMs) that answer questions about web video.
The evaluation recipe has been transplanted with them, which amounts to sampling frames or views, posing a multiple-choice question and reporting accuracy~\cite{zhao2025urbanvideobench,zha2026aircopbench}.
Aerial and remote sensing (RS) video, however, differs from web video in the variables that matter most for perception.
The camera moves in six degrees of freedom and its altitude sets the ground sampling distance; objects of interest are often a few pixels wide; views from several platforms must be related through geometry rather than appearance; and the relevant time scale ranges from tens of frames per second on a drone to months between satellite passes.
A benchmark that ignores these variables can report steady progress, while the capabilities an operator needs remain untested.

\para{Position.}
\emph{The remaining gap in aerial and remote sensing video understanding is a gap in geometry and time, not in appearance or scale, and current benchmarks are not built to measure it.}
The paper argues this position based on a wide set of published results.
In particular, it investigates what the reported results of 67 literature works already imply, when they are read with within-paper contrasts, and normalized against each paper's own chance and human baselines.

\para{Convergent meta-analysis.}
The aerial MLLM literature is recent, while the aerial perception one is extensive but predates MLLMs.
Neither corpus alone can carry a position.
The analysis therefore separates the evidence into four tiers (\cref{sec:method}) and adopts a convergence rule, under which a claim becomes a position only when peer-reviewed and preprint evidence point the same way; claims supported only by preprints are reported as tentative, and claims supported only by pre-MLLM studies are reported as established for classical perception with the transfer to MLLMs left open.

\para{Contributions.}
\begin{itemize}
\item A facet-coded, tiered meta-analysis of aerial and RS video understanding, comprising human-normalized scores for 112 tasks and 240 zero-shot model rows in 10 MLLM benchmarks, 21 retention contrasts over 333 method or configuration evaluations from 15 multi-method perception papers, a model-matched aerial versus ground comparison, and a reporting audit of 20 aerial benchmarks (\cref{sec:method,sec:findings}).
\item Eleven claims graded by the convergence rule, of which six are convergent positions, one rests on peer-reviewed closed-loop evidence alone, one is an established pre-MLLM weakness, and three are tentative, descriptive or divergent (\cref{sec:findings}).
\item Five principles derived from the convergent claims, each stated with the evidence that would falsify it (\cref{sec:position}).
\item The \textbf{AVU-Card}, a reporting card for aerial video benchmarks, with a worked example (\cref{sec:card}).
\end{itemize}

\section{What Makes Aerial Video Different}
\label{sec:background}

\para{Two axes of ``multi-facet''.}
Aerial and RS video understanding is multi-facet in two senses, and the analysis uses both.
The first axis is \emph{what is asked}.
Every benchmark task is coded to one of seven competence facets, namely perception (recognition, attributes, scene state), counting, spatial reasoning (relations, distance, direction, grounding), temporal reasoning (order, change, events, prediction), multi-view association (across drones, or between air and ground), ego-motion or self-state (the camera's own pose, motion and altitude), and planning (what to do next).
The second axis is \emph{under which acquisition conditions}, which range over altitude and ground sampling distance, view angle, object size, time horizon and frame rate, illumination and sensor, and the number of cooperating platforms.
Web-video benchmarks vary the first axis and hold the second nearly constant.
Aerial video varies both and the second axis is where aerial footage departs from the data on which MLLMs are pre-trained.

\para{The temporal range is wider than in web video.}
Drone video runs at 10--60\,fps; satellite video at a few to 30\,fps over tens of seconds (VISO uses 10\,fps clips~\cite{yin2022viso}); and RS time series revisit a site after days, months or years~\cite{irvin2025teochat,xuan2025dynamicvl}.
``Video understanding'' in this domain therefore spans frame-to-frame motion, event recognition over seconds~\cite{mou2020era}, and change over seasons.
Each aerial MLLM benchmark in the corpus covers essentially one of these scales, for instance a single instant in AirCopBench~\cite{zha2026aircopbench} and flight clips of 10\,s to 10\,min in UrbanVideo-Bench~\cite{zhao2025urbanvideobench}.

\para{Related work.}
Aerial perception has a decade of multi-method benchmarks for detection, tracking, counting, re-identification and geo-localization~\cite{mueller2016uav123,li2017dtb70,Du2018UAVDT,Zhu2022TPAMI,wen2021dronecrowd,zhang2022vtuav,cheng2023soda,ding2022dota,zheng2020university1652,zhu2023sues200,nguyen2025agvpreid}.
Language-grounded aerial benchmarks followed, with embodied navigation and question answering~\cite{liu2023aerialvln,lee2025citynav,wang2025travel,xiao2025uavon,gao2026openfly,zhao2025cityeqa,yao2026aeroverse}, and more recently MLLM question answering over drone video and multi-drone views~\cite{zhao2025urbanvideobench,zha2026aircopbench}, and referring and grounding~\cite{sun2026refdrone,chen2026aerialmind}, followed by a wave of preprints~\cite{dai2025mmuavbench,zhang2026spatialuav,yi2026aeroground,liu2026uavdualcog,sun2026uavreason,zhan2026uavbench,sun2026droneeyes,weng2025uavde}.
Temporal RS benchmarks evaluate change and multi-temporal reasoning~\cite{danish2025geobench,wang2025xlrs,soni2025earthdial,irvin2025teochat,wang2025disasterm3,xuan2025dynamicvl}.
On the ground-view side, spatial-intelligence benchmarks~\cite{yang2025vsibench,zhou2025vlm4d,wu2026spatialscore,yang2026mmsibench,gholami2026ego3d,zhang2025spar} probe metric and relational reasoning, and diagnostic studies show that many ``video'' questions can be answered from a single frame~\cite{buch2022revisiting,cores2025lostintime} or a short clip~\cite{mangalam2023egoschema}, which recent benchmarks counter with text-only filtering or blind baselines~\cite{fu2025videomme,niu2025ovobench,lin2024streamingbench}.
This paper is, to the authors' knowledge, the first to pool these literatures quantitatively for the aerial domain.
It follows the spirit of documentation proposals for datasets and models~\cite{gebru2021datasheets,mitchell2019modelcards}, but asks a narrower question, namely what an aerial video benchmark must report for its accuracy numbers to support the claims made from them.

\section{Meta-Analysis Method}
\label{sec:method}

\begin{table*}[t]
\centering
\scriptsize
\setlength{\tabcolsep}{3.0pt}
\setlength{\aboverulesep}{0pt}\setlength{\belowrulesep}{0pt}
\renewcommand{\arraystretch}{1.15}
\rowcolors{2}{gray!20}{gray!2}
\renewcommand{\tabularxcolumn}[1]{m{#1}}
\begin{tabularx}{\textwidth}{>{\raggedright\arraybackslash}m{2.3cm} >{\raggedright\arraybackslash}X >{\centering\arraybackslash}m{3.2cm} >{\centering\arraybackslash}m{2.3cm}}
\toprule
\rowcolor{gray!50} \multicolumn{1}{c}{\textbf{Tier}} & \multicolumn{1}{c}{\textbf{Papers (venue)}} & \textbf{Extracted} & \textbf{Normalization} \\
\midrule
\textbf{T1} Aerial language-grounded and closed-loop benchmarks (peer-reviewed) &
AirCopBench (AAAI'26)~\cite{zha2026aircopbench}, UrbanVideo-Bench (ACL'25)~\cite{zhao2025urbanvideobench}, AerialMind (AAAI'26)~\cite{chen2026aerialmind}, RefDrone (CVPR'26 F)~\cite{sun2026refdrone}, Sky2Ground (CVPR'26)~\cite{wang2026sky2ground}, AeroVerse (TPAMI'26)~\cite{yao2026aeroverse}, CityEQA (EMNLP'25)~\cite{zhao2025cityeqa}, AerialVLN (ICCV'23)~\cite{liu2023aerialvln}, CityNav (ICCV'25)~\cite{lee2025citynav}, TravelUAV (ICLR'25)~\cite{wang2025travel}, UAV-ON (MM'25)~\cite{xiao2025uavon}, OpenFly (ICLR'26)~\cite{gao2026openfly} &
Per-task tables (24 zero-shot rows on AirCopBench), scale pairs, fine-tuning, closed-loop success, audit & HNS; chance-normalized if no human \\
\textbf{T2} Multi-method aerial perception, 2016--2025 (peer-reviewed) &
UAV123~\cite{mueller2016uav123}, DTB70~\cite{li2017dtb70}, UAVDT~\cite{Du2018UAVDT}, NDFT~\cite{Wu2019NDFT}, VisDrone DET/VID/SOT/MOT reports and TPAMI~\cite{Du2019VisDroneDET,Cao2021VisDroneDET,Zhu2019VisDroneVID,Du2019VisDroneSOT,Wen2019VisDroneMOT,Chen2021VisDroneMOT,Zhu2022TPAMI}, SAHI~\cite{Akyon2022SAHI}, Okutama~\cite{barekatain2017okutama}, University-1652~\cite{zheng2020university1652}, ERA~\cite{mou2020era}, DroneCrowd~\cite{wen2021dronecrowd}, UAV-Human~\cite{li2021uavhuman}, VTUAV~\cite{zhang2022vtuav}, DroneVehicle~\cite{sun2022dronevehicle}, VISO~\cite{yin2022viso}, DOTA~\cite{ding2022dota}, SODA~\cite{cheng2023soda}, SUES-200~\cite{zhu2023sues200}, SeaDronesSee~\cite{varga2022seadronessee}, MAVREC~\cite{dutta2024mavrec}, AG-VPReID~\cite{nguyen2025agvpreid}, Video2BEV~\cite{ju2025video2bev}, CST Anti-UAV~\cite{xie2025cstantiuav} &
Per-method scores under two acquisition conditions (5--41 methods per contrast) & Retention $=$ harder / easier, per method; median \\
\textbf{T3} Ground and temporal-RS analogues (peer-reviewed) &
Spatial benchmarks VSI-Bench~\cite{yang2025vsibench}, VLM4D~\cite{zhou2025vlm4d}, SpatialScore~\cite{wu2026spatialscore}, MMSI-Bench~\cite{yang2026mmsibench}, Ego3D-Bench~\cite{gholami2026ego3d}, SPAR-Bench~\cite{zhang2025spar}; temporal RS benchmarks GEOBench-VLM~\cite{danish2025geobench}, XLRS-Bench~\cite{wang2025xlrs}, EarthDial~\cite{soni2025earthdial}, TEOChat~\cite{irvin2025teochat}, DisasterM3~\cite{wang2025disasterm3}, DynamicVL~\cite{xuan2025dynamicvl}; video anchors ATP~\cite{buch2022revisiting}, EgoSchema~\cite{mangalam2023egoschema}, Video-MME~\cite{fu2025videomme}, OVO-Bench~\cite{niu2025ovobench}, StreamingBench~\cite{lin2024streamingbench}, TVBench~\cite{cores2025lostintime}; forensic search ForeSea~\cite{park2026foresea} &
Per-task human, chance and blind rows (166 zero-shot rows on the six spatial benchmarks); single- vs multi-temporal and short vs long contrasts & HNS; retention \\
\textbf{T4} Preprints 2025--26 (\pre) &
MM-UAVBench\pre~\cite{dai2025mmuavbench}, SpatialUAV\pre~\cite{zhang2026spatialuav}, AeroGround\pre~\cite{yi2026aeroground}, UAV-DualCog\pre~\cite{liu2026uavdualcog}, UAVReason\pre~\cite{sun2026uavreason}, UAVBench\pre~\cite{zhan2026uavbench}, DroneEyes\pre~\cite{sun2026droneeyes}, UAVDE-2M\pre~\cite{weng2025uavde} &
Same as T1 (50 zero-shot rows on the three HNS benchmarks) & HNS; used for corroboration only \\
\bottomrule
\end{tabularx}
\caption{\textbf{Evidence tiers.} 67 papers. Numbers are read from the papers' own tables; values read from bar charts (UAVDT tracking) or plot legends (VisDrone-SOT19, VTUAV, UAV123) are used only in retention contrasts, never in HNS pooling. Tiers are pooled separately and compared by direction, never averaged together. \pre\ marks preprints.}
\label{tab:corpus}
\end{table*}

\para{Corpus and tiers.}
The corpus (\cref{tab:corpus}) was assembled up to September 2026 from literature searches for aerial, remote sensing and ground-view benchmarks, and multi-method studies that report per-method or per-model numbers.
A paper is included when it reports numbers relevant to at least one analysis below, so the corpus is a curated sample of this evidence rather than a census.
Venue status was checked against the published record of each paper and papers without a verifiable peer-reviewed version are treated as preprints.
Values were transcribed from tables, and values from tables available only as images were checked against numbers quoted in the paper's text where such numbers exist.

\para{Human-normalized score.}
For a task $t$ with the paper's own chance level $r_t$ and human accuracy $h_t$, a model score $s_t$ is normalized as
\begin{equation}
\hns_t = \min\!\left(1,\ \frac{s_t - r_t}{h_t - r_t}\right),
\end{equation}
so that 0 is chance and 1 is human level; negative values (below chance) are kept.
Only zero-shot rows enter the pooled analyses; fine-tuned and subset-evaluated rows are used in contrasts only.
Tasks with $h_t - r_t < 20$ points are excluded as unstable (7 of 119 tasks); for numeric-answer tasks without a stated chance level, the paper's frequency-based chance baseline is used as $r_t$.
Each task is coded to one competence facet (\cref{sec:background}); a model's facet score is the mean HNS over that facet's tasks, and the facet value reported for a benchmark is the median over its models, and for a tier the median over the model rows of its benchmarks (the ``median model''). For example, AirCopBench's observing-posture task is coded as ego-motion, object matching across drones as multi-view association, and the decisions on when, what, who and why to collaborate as planning.

\para{Retention contrasts.}
For the multi-method papers of T2, the analysis extracts every table in which the same methods are evaluated under an easier and a harder acquisition condition, and computes the per-method retention (score under the harder condition divided by score under the easier one).
The median over methods and the number of methods that lose accuracy are reported; contrasts that go the other way are kept as counter-examples.

\para{Model-matched comparison.}
To test whether a weakness is specific to aerial data, each model evaluated both on an aerial benchmark (T1 or T4) and on at least one ground spatial benchmark (T3) is compared through its mean HNS over all tasks of each benchmark.

\para{Convergence rule.}
Tiers are pooled separately and compared by the direction of an effect, not by its magnitude.
A claim is \emph{convergent} when peer-reviewed evidence (T1--T3) and preprint evidence (T4) agree; \emph{tentative} when only T4 supports it; \emph{established, pre-MLLM} when only T2 supports it; and \emph{peer-reviewed only} when T1--T3 support it and no preprint tests it.
Sensitivity is assessed by best-model versus median-model scores and by leaving one benchmark out of each tier.

\section{Findings}
\label{sec:findings}

Findings are stated as claims K1--K11 with their evidence by tier; \cref{tab:ledger} in \cref{sec:position} summarizes their status.

\begin{figure}[t]
\centering
\includegraphics[width=\linewidth]{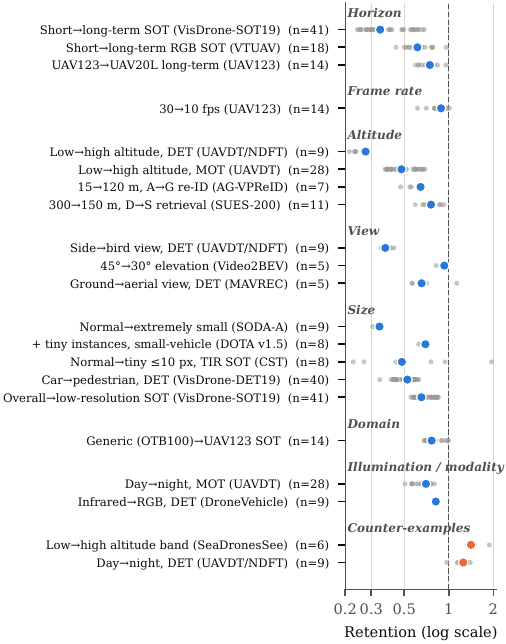}
\caption{\textbf{Acquisition conditions control aerial perception (T2).} Retention is the score under the harder condition divided by the score under the easier one. Each row is a within-paper contrast; gray dots are individual methods and the blue dot is the median retention. Across the 19 degrading contrasts, 315 of 318 method or configuration evaluations lose accuracy. Orange rows are the two counter-examples in the corpus. UAVDT MOT values are read from bar charts and the VisDrone-SOT19, VTUAV and UAV123 values from plot legends.}
\label{fig:retention}
\end{figure}

\subsection{Scale, appearance and geometry}

\begin{figure*}[t]
\centering
\includegraphics[width=\textwidth]{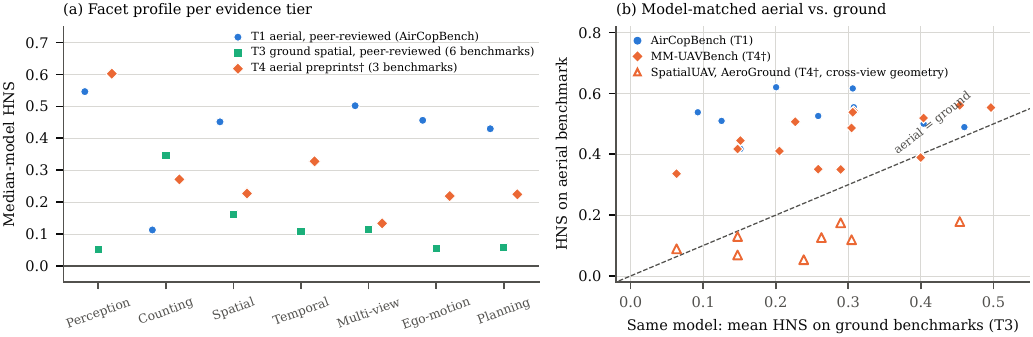}
\caption{\textbf{(a)} Median-model HNS per competence facet and tier (T1 with 24 models and 13 tasks, T3 with 166 model rows over 6 benchmarks and 56 tasks, T4 with 50 model rows over 3 benchmarks and 43 tasks). T3 perception rests on two tasks. \textbf{(b)} Each point is one model, plotting its HNS on an aerial benchmark against its mean HNS on the ground spatial benchmarks where it is also evaluated. Models score \emph{higher} on AirCopBench (9/9) and MM-UAVBench\pre\ (13/14) than on ground spatial benchmarks, and \emph{lower} on the geometry-heavy SpatialUAV\pre\ and AeroGround\pre\ (7 of 8 model--benchmark pairs).}
\label{fig:facets}
\end{figure*}

\para{K1. Perception leads the aerial facet profile (convergent, sensitive).}
On AirCopBench (T1, 24 models, 13 tasks) the median model reaches HNS 0.55 on perception against 0.11--0.50 on the other facets; in the pooled preprints (T4) the figures are 0.60 against 0.13--0.33 (\cref{fig:facets}a).
The lead is a property of the typical model and is sensitive in both tiers, since for the best model per facet, AirCopBench perception (0.70) is no longer ahead of spatial, multi-view, planning and ego-motion (0.73--1.00, each resting on one to three tasks), and the T4 lead rests on MM-UAVBench\pre, without which perception falls to 0.15, below counting (0.24), spatial (0.23) and ego-motion (0.22).
Beyond ``perception leads'', the facet rankings diverge across tiers (counting is the weakest facet in T1 at 0.11 but mid-table in T4 at 0.27; multi-view is the weakest in T4 at 0.13 but among the strongest in T1 at 0.50), so no finer ranking is claimed.

\begin{figure}[t]
\centering
\includegraphics[width=\linewidth]{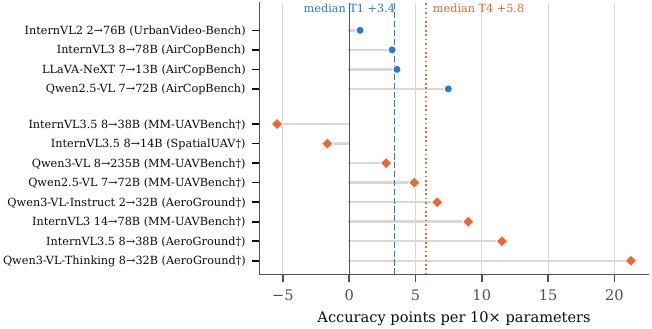}
\caption{\textbf{Within-family scale slopes} on aerial benchmarks (overall accuracy; $\dagger$ preprint; preprint family--benchmark contrasts from the three HNS benchmarks). Both tiers give a few points per tenfold increase in parameters, against human gaps of 19--52 points.}
\label{fig:scale}
\end{figure}

\para{K2. Parameter scale is not the lever (convergent).}
Within model families evaluated at two or more sizes (\cref{fig:scale}), overall accuracy rises by a median of $+3.4$ points per tenfold increase in parameters on peer-reviewed aerial benchmarks (4 families; $+0.8$ to $+7.5$), and by $+5.8$ on the three preprint HNS benchmarks (8 family--benchmark contrasts; $-5.4$ to $+21.3$, 2 negative).
InternVL2 moves from 27.6 to 28.9 between 2B and 76B on UrbanVideo-Bench~\cite{zhao2025urbanvideobench}.
At $+3.4$ points per decade, closing the 19.1-point gap between the best open model and humans on AirCopBench would take about 5.6 decades of parameters; the preprint gaps are 25.0, 39.5 and 51.7 points~\cite{dai2025mmuavbench,zhang2026spatialuav,yi2026aeroground}.
Domain data is more effective but uneven, as supervised fine-tuning lifts Qwen2.5-VL-7B on AirCopBench from 47.3 to 74.3 overall and from 25.9 to 75.3 on counting, yet only from 25.9 to 33.3 on the observer-posture task, and simulation-to-real training leaves posture on the small real-image test subset unchanged at 11.1~\cite{zha2026aircopbench}.
On UrbanVideo-Bench, simulation-to-real fine-tuning adds 3.2--5.2 points; LoRA on AeroGround\pre\ adds 1.5--2.2, and models tuned for ground spatial reasoning trail the best generalists (29.7 vs.\ 56.7 on SpatialUAV\pre, 33.8 vs.\ 54.4 on AeroGround\pre).

\para{K3. Acquisition geometry controls perception (convergent).}
\Cref{fig:retention} collects the T2 contrasts.
Moving from low to high altitude retains a median 0.27 of detection AP on UAVDT (9/9 configurations drop~\cite{Wu2019NDFT}) and 0.65 of re-identification rank-1 between 15 and 120\,m (7/7 variants~\cite{nguyen2025agvpreid}); for drone-to-satellite retrieval the harder direction is down, and flying at 150 instead of 300\,m retains 0.76 of recall (11/11 variants~\cite{zhu2023sues200}), so the relevant variable is the match between altitude and task rather than altitude as such.
A bird's-eye view retains 0.37 of side-view detection AP on UAVDT~\cite{Wu2019NDFT}; extremely small objects retain 0.34 of normal-size AP on SODA-A (9/9)~\cite{cheng2023soda}; and adding sub-10-pixel instances to the same DOTA images cuts small-vehicle AP to 0.70 of its v1.0 value (8/8) while the other 14 classes keep 0.97~\cite{ding2022dota}.
Geometry-aware inputs help in most tests (\cref{tab:geoinputs}), from flight metadata and view-angle experts to maps, bird's-eye-view representations and ground-view pre-training.
The gains are smallest, and sometimes negative, where the geometry is hardest, as full metadata adds only 1.9 AP at high altitude and altitude-only disentanglement lowers it from 18.7 to 15.7~\cite{Wu2019NDFT}, altitude experts add 0.8 AP50 against 3.1 for view-angle experts~\cite{varga2022seadronessee}, and homography compensation hurts all 8 trackers under out-of-plane rotation~\cite{li2017dtb70}.
Two counter-examples show that altitude is an interacting variable, not a fixed penalty, since on SeaDronesSee the highest altitude band is the easiest (median 1.41), where altitude is confounded with view angle and class mix, and detection on UAVDT is easier at night (1.25)~\cite{Wu2019NDFT} while tracking at night is harder (0.70)~\cite{Du2018UAVDT}.
Preprints point the same way, with frontier MLLMs reaching 0.95--6.0 J\&F on tiny referring targets against 33.1 for a segmentation model, partly because their boxes are scored with mask metrics~\cite{sun2026droneeyes}, and an open-vocabulary detector keeping 0.13--0.19 of its COCO accuracy on VisDrone~\cite{weng2025uavde}.

\begin{table}[t]
\centering
\scriptsize
\setlength{\tabcolsep}{2.4pt}
\setlength{\aboverulesep}{0pt}\setlength{\belowrulesep}{0pt}
\renewcommand{\arraystretch}{1.1}
\rowcolors{2}{gray!20}{gray!2}
\renewcommand{\tabularxcolumn}[1]{m{#1}}
\begin{tabularx}{\linewidth}{>{\raggedright\arraybackslash}X >{\centering\arraybackslash}m{2.35cm} >{\centering\arraybackslash}m{1.65cm}}
\toprule
\rowcolor{gray!50} \multicolumn{1}{c}{\textbf{Geometry-aware input}} & \textbf{Effect} & \textbf{Source} \\
\midrule
Altitude, view, weather metadata (training) & +12.0 AP night, +11.5 front view, +1.9 high altitude & NDFT~\cite{Wu2019NDFT} \\
View-angle experts & +3.1 AP50 (average) & SeaDrones\-See~\cite{varga2022seadronessee} \\
Camera-motion compensation (homography) & 8/8 trackers improve overall & DTB70~\cite{li2017dtb70} \\
Geographic semantic map (GNSS + map) & Success $\times$2.5--5.7 (test-unseen) & CityNav~\cite{lee2025citynav} \\
Bird's-eye-view (3DGS) representation & +4.4 retrieval AP & Video2BEV~\cite{ju2025video2bev} \\
Same-region ground-view pre-training & AP 33.2 $\rightarrow$ 44.8 & MAVREC~\cite{dutta2024mavrec} \\
Estimated / ground-truth 3D map (ground, T3) & 43.1 $\rightarrow$ 56.0 / 79.4 & Ego3D~\cite{gholami2026ego3d} \\
\bottomrule
\end{tabularx}
\caption{\textbf{Geometry-aware inputs and training tested in the corpus} (exceptions in the text). Among the aerial question-answering benchmarks, only AeroVerse gives pose to the evaluated models (\cref{tab:audit}).}
\label{tab:geoinputs}
\end{table}

\para{K4. Aerial MLLM benchmarks differ by whether they test geometry (divergent, informative).}
If aerial deficits were specific to aerial imagery, the same model should score lower on aerial than on ground benchmarks.
\Cref{fig:facets}b shows the opposite for the peer-reviewed benchmark, as all 9 models shared between AirCopBench and the ground spatial benchmarks score higher on AirCopBench (median $+0.27$ HNS), and 13 of 14 score higher on MM-UAVBench\pre\ ($+0.19$).
The only aerial benchmarks on which the same models score lower than on ground are the two preprints built around cross-view metric geometry, SpatialUAV\pre\ and AeroGround\pre\ (7 of 8 model--benchmark pairs; median $-0.08$ and $-0.15$).
The ground benchmarks themselves are hard where geometry is explicit, since the lowest ground facets are ego-motion and perception (0.05 each, perception on two tasks only), GPT-4o without images scores below chance on MMSI-Bench (22.7 vs.\ 25)~\cite{yang2026mmsibench}, and on SpatialScore a blind GPT-5 is within 2.3 points of chance while the sighted model reaches 58.1~\cite{wu2026spatialscore}.
The divergence is therefore informative, since the difficulty of an aerial benchmark tracks how much explicit geometry it demands.

\subsection{Time, action and evaluation}

\para{K5. Longer horizons and temporal change cause large losses (convergent).}
\Cref{tab:time} collects the temporal contrasts.
Long horizons are the largest loss for classical aerial trackers (retention 0.34 across 41 trackers), frame-rate reduction hurts 13 of 14 trackers, and temporal modeling brings little in uncontrolled comparisons, given that in the VisDrone video challenge the winner is a per-frame detector, although among the organizers' baselines the video detectors lead (17.7 vs.\ 15.9 AP)~\cite{Zhu2019VisDroneVID}, and that on ERA the median video model reaches 0.87 of the median frame model~\cite{mou2020era}.
For MLLMs on RS time series, accuracy drops sharply from single-image to multi-temporal or change tasks in most contrasts of \cref{tab:time} (counting on XLRS-Bench and event types on EarthDial's xBD set are exceptions, at 0.98 and 0.79~\cite{soni2025earthdial}), and GPT-4o falls below chance on XLRS-Bench change counting~\cite{wang2025xlrs}.
More temporal context helps when it is given, as on TEOChat's QFabric task all 5 models gain from 2 to 5 images (+1.0 to +8.9 F1), most of all the temporally trained ones (+7.6 and +8.9)~\cite{irvin2025teochat}.
Ground-video anchors explain why accuracy alone cannot settle the question, since shuffled frames keep 0.98 of the accuracy on MVBench and 0.92 on TVBench (but half of its margin above chance)~\cite{cores2025lostintime}, and one well-chosen frame matches the best video model~\cite{buch2022revisiting}.
In aerial MLLM benchmarks the temporal facet ranks second in T4 (median HNS 0.33, but 0.04 without MM-UAVBench\pre), the questions use sampled or key frames, and none is ablated for frame count or order (K7), so the score does not show that they are temporal; where order is tested directly, AeroGround\pre's reverse-sequence task has a median HNS of 0.04.

\begin{table}[t]
\centering
\scriptsize
\setlength{\tabcolsep}{2.2pt}
\setlength{\aboverulesep}{0pt}\setlength{\belowrulesep}{0pt}
\renewcommand{\arraystretch}{1.08}
\rowcolors{2}{gray!20}{gray!2}
\renewcommand{\tabularxcolumn}[1]{m{#1}}
\begin{tabularx}{\linewidth}{c >{\raggedright\arraybackslash}X >{\centering\arraybackslash}m{1.75cm} >{\centering\arraybackslash}m{1.45cm}}
\toprule
\rowcolor{gray!50} \textbf{Tier} & \multicolumn{1}{c}{\textbf{Temporal contrast}} & \textbf{Result} & \textbf{Source} \\
\midrule
T2 & Short $\rightarrow$ long-term SOT, 41 trackers & 0.34 (41/41) & \cite{Du2019VisDroneSOT} \\
T2 & Short $\rightarrow$ long-term RGB SOT, 18 trackers & 0.61 (18/18) & \cite{zhang2022vtuav} \\
T2 & 30 $\rightarrow$ 10\,fps, 14 trackers & 0.89 (13/14) & \cite{mueller2016uav123} \\
T2 & Temporal vs.\ per-frame entries, mean (organizers' baselines) & 15.2 vs.\ 19.5 (17.7 vs.\ 15.9) AP & \cite{Zhu2019VisDroneVID} \\
T2 & Median video vs.\ median frame model (events) & 0.87 & \cite{mou2020era} \\
T3 & Single image (mean of 3 sets) $\rightarrow$ multi-temporal fMoW, GPT-4o & 84.9 $\rightarrow$ 21.4 & \cite{soni2025earthdial} \\
T3 & Scene understanding $\rightarrow$ temporal tasks & 0.58 (4/4) & \cite{danish2025geobench} \\
T3 & Anomaly reasoning $\rightarrow$ change counting & 0.50 (8/8) & \cite{wang2025xlrs} \\
T3 & Single-choice $\rightarrow$ select-all change analysis & 0.27 (16/16) & \cite{xuan2025dynamicvl} \\
T3 & Short $\rightarrow$ long web video & 0.76 (16/16) & \cite{fu2025videomme} \\
T3 & Single chosen frame vs.\ best video model & 1.00 & \cite{buch2022revisiting} \\
T3 & Ordered $\rightarrow$ shuffled frames, TVBench (acc.\ / margin) & 0.92 / 0.51 & \cite{cores2025lostintime} \\
T4 & Reverse-sequence task, median model HNS & 0.04 & \cite{yi2026aeroground}\pre \\
\bottomrule
\end{tabularx}
\caption{\textbf{Temporal evidence by tier.} Unless stated otherwise, values are retention (harder / easier condition, median over methods; methods that drop in brackets). Margin denotes accuracy above chance.}
\label{tab:time}
\end{table}

\para{K6. Competence collapses from answering to acting (peer-reviewed, T1).}
Closed-loop aerial navigation reaches HNS 0.020 on AerialVLN (1.6\% vs.\ 80.8\% success for humans)~\cite{liu2023aerialvln}, 0.076 on CityNav with a map and 0.020 without~\cite{lee2025citynav}, and 0.18 on TravelUAV (17.5\% vs.\ 94.5\%, with an oracle assistant)~\cite{wang2025travel}; chance is taken as 0 where no random agent is reported.
Generalization compounds the gap, with seen-to-unseen success retention of 0.57 on AerialVLN (7/8 methods), 0.65 on CityNav (humans 0.99) and 0.26 for unseen maps on TravelUAV (3/3).
On CityEQA, an MLLM given the right image answers at 4.37 on a 1--5 scale, but the full agent reaches 3.00 against 4.94 for humans, so the loss lies in exploration, not in answering~\cite{zhao2025cityeqa}.

\para{K7. Aerial benchmarks cannot yet test their own claims (convergent).}
\Cref{tab:audit} audits the 12 peer-reviewed T1 benchmarks and the 8 aerial preprints.
Among the peer-reviewed benchmarks, only AirCopBench reports per-task human and random baselines, and it uses still images; three preprints do so as well, none of them with video input for all tasks.
No aerial benchmark reports variance across runs; blind scores appear only in two closed-loop papers, none of the aerial question-answering benchmarks reports a blind or single-frame score, and 3 of them (AirCopBench, UrbanVideo-Bench and SpatialUAV\pre) use blind models only to filter questions, leaving no blind floor against which accuracy can be read.
At least twelve of the 20 papers contain a verified inconsistency between text and tables, an aggregate that does not reproduce, or overlapping data splits.
No peer-reviewed aerial benchmark in the corpus therefore combines video input with per-task human and random baselines, which is itself the strongest reporting finding.
Accuracy also hides whether an answer rests on the right evidence.
On ForeSeaQA, a forensic-search benchmark of ground surveillance video, eleven video MLLMs and retrieval methods answer 36.2--66.0\% of the questions correctly but localize the supporting interval with a temporal IoU of only 3.5--13.6\%, and ground-truth captions raise accuracy to 83.2\% but IoU only to 31.1\%~\cite{park2026foresea}.

\begin{table}[t]
\centering
\scriptsize
\setlength{\tabcolsep}{2.4pt}
\setlength{\aboverulesep}{0pt}\setlength{\belowrulesep}{0pt}
\rowcolors{2}{gray!20}{gray!2}
\begin{tabular}{l c c}
\toprule
\rowcolor{gray!50} \multicolumn{1}{c}{\textbf{Reported item}} & \textbf{T1 (12)} & \textbf{T4\pre\ (8)} \\
\midrule
Human baseline, any / per task & 5 / 1 & 4 / 3 \\
Random or chance baseline & 7 & 3 \\
Variance across runs or seeds & 0 & 0 \\
Blind (text-only) score & 2 & 0 \\
Frame-count, clip-length or view-count ablation & 4 & 1 \\
Video as input for all tasks & 2 & 1 \\
Telemetry or pose given to the model & 5 & 0 \\
\quad of which QA/perception benchmarks & 1 of 6 & 0 of 8 \\
$\geq$1 verified internal inconsistency & 9 & 3 \\
Video + per-task human + random & 0 & 0 \\
\bottomrule
\end{tabular}
\caption{\textbf{Reporting audit} of the aerial benchmarks in T1 and T4 (counts of benchmarks). The two blind scores come from closed-loop papers (CityEQA, AerialVLN); no aerial question-answering benchmark reports one.}
\label{tab:audit}
\end{table}

\para{K8. Views need geometry (convergent in weak form).}
Relating views by appearance fails when viewpoints differ, as ground-view queries reach 1.20 recall@1 on University-1652 against 58.5 for drone queries of the same buildings~\cite{zheng2020university1652}, and fine-tuning a feed-forward geometry model on ground, aerial and satellite views together lowers its rotation accuracy relative to the zero-shot model from 73.6 to 55.4 overall and from 79.2 to 29.7 on aerial images, while zero-shot it recovers no satellite camera position at all~\cite{wang2026sky2ground}.
In MM-UAVBench\pre, adding a view lowers accuracy below the best single view in 15 of 20 model--setting pairs (median $-5.9$ for air--ground)~\cite{dai2025mmuavbench}.
The weak form (``views help only through geometry'') is convergent; the strong form (``views hurt'') remains tentative.

\para{K9--K11.}
\emph{Ego-motion} (K9, tentative for aerial data) is, with perception, the weakest ground facet in T3, and self-referenced questions trail world-referenced ones in accuracy in 35 of 36 models on UAV-DualCog\pre, although box overlap is higher for the self-referenced questions in 26 of 36 models~\cite{liu2026uavdualcog}; motion errors, such as reading a gimbal tilt as vertical motion, make up 58--62\% of the sampled navigation and recall errors on UrbanVideo-Bench~\cite{zhao2025urbanvideobench}; but the single AirCopBench posture task sits mid-table.
The \emph{human gap} (K10) is reported per tier, amounting to 19.1 points on AirCopBench and 25.0--51.7 on the preprints.
\emph{Persistence} (K11, established pre-MLLM) is shown by VisDrone, where on the 2021 test set the winner reaches 39.4 AP against 37.4 for the 2019 winner, yet all 40 detectors of 2019 and all 8 trackers of 2021 remain weaker on pedestrians than on cars~\cite{Du2019VisDroneDET,Cao2021VisDroneDET,Chen2021VisDroneMOT}.

\para{Robustness.}
Lowering the headroom threshold from 20 to 10 or 0 points (116 or all 119 tasks kept), raising it to 30 points (106 tasks), or removing the cap at human level leaves perception as the leading facet in T1 and T4, and ego-motion and perception as the two weakest facets in T3.
Leaving one benchmark out is more demanding, since perception stays first in T4 only when both MM-UAVBench\pre\ and AeroGround\pre\ are kept (without AeroGround\pre, counting leads with 0.71 against 0.60), which is why K1 is graded as sensitive.
In T3, ego-motion and perception remain among the three weakest facets in five of the six leave-one-out runs, the exception being the removal of MMSI-Bench, after which perception rests on a single VLM4D task.
The model-matched comparison keeps its direction under every threshold, as 8 or 9 of the 9 AirCopBench models score higher than on ground benchmarks and 7 of 8 cross-view preprint pairs score lower.
Finally, the 11 degrading retention contrasts taken only from printed tables still show 118 of 120 evaluations losing accuracy, and leaving out any one of the four peer-reviewed scale contrasts keeps the median slope between $+3.2$ and $+3.6$ points per decade.

\section{From Evidence to Position}
\label{sec:position}

\begin{table}[t]
\centering
\scriptsize
\setlength{\tabcolsep}{2.2pt}
\setlength{\aboverulesep}{0pt}\setlength{\belowrulesep}{0pt}
\renewcommand{\arraystretch}{1.12}
\rowcolors{2}{gray!20}{gray!2}
\renewcommand{\tabularxcolumn}[1]{m{#1}}
\begin{tabularx}{\linewidth}{l >{\raggedright\arraybackslash}X >{\centering\arraybackslash}m{0.55cm} >{\centering\arraybackslash}m{0.55cm} >{\centering\arraybackslash}m{1.35cm} >{\centering\arraybackslash}m{0.75cm}}
\toprule
\rowcolor{gray!50} & \multicolumn{1}{c}{\textbf{Claim}} & \textbf{T1--3} & \textbf{T4} & \textbf{Status} & \textbf{$\rightarrow$ P} \\
\midrule
K1 & Perception leads the facet profile & \cmark & \cmark & Convergent (LOO-sensitive) & P2 \\
K2 & Parameter scale is not the lever & \cmark & \cmark & Convergent & P1 \\
K3 & Acquisition geometry controls accuracy; geometry inputs help & \cmark & \cmark & Convergent & P1, P2 \\
K4 & Aerial MLLM benchmarks under-test geometry & \cmark & \cmark & Divergent, informative & P1, P5 \\
K5 & Longer horizons and temporal change cause large losses & \cmark & \cmark & Convergent & P3 \\
K6 & Answering does not transfer to acting & \cmark & -- & Peer-reviewed only & P3, P5 \\
K7 & Benchmarks cannot test their own claims & \cmark & \cmark & Convergent & P5 \\
K8 & Views help only through geometry & \cmark & \cmark & Convergent (weak form) & P1, P4 \\
K9 & Ego-motion is among the weakest facets & \pmark & \cmark & Tentative & P1, P4 \\
K10 & Human gap $\geq$19 points & \cmark & \cmark & Per tier & P5 \\
K11 & Weaknesses persist across a decade & \cmark & -- & Established, pre-MLLM & P2, P3 \\
\bottomrule
\end{tabularx}
\caption{\textbf{Claim ledger.} \cmark\ supported; \pmark\ supported only by ground-view analogues (T3); -- not tested in that tier. LOO denotes leave-one-benchmark-out. The last column maps claims to the principles of \cref{sec:position}.}
\label{tab:ledger}
\end{table}

The convergent claims support five principles.
Each is stated with the observation that would falsify it, so that the position can be tested by the benchmarks it calls for.

\para{P1. Geometry as an input (K2, K3, K4, K8, K9).}
Aerial models should receive the camera's pose, altitude and ground sampling distance as inputs, and aerial benchmarks should provide them.
Most geometry-aware inputs in the corpus improve accuracy, from flight metadata to maps and bird's-eye-view representations, whereas parameter scale moves overall accuracy little and fine-tuning leaves the one peer-reviewed self-state task almost unchanged (+7 points against +49 for counting).
Among the aerial question-answering benchmarks only AeroVerse provides pose to the model, although several record camera poses to generate their ground truth~\cite{zhang2026spatialuav,yi2026aeroground}.
The principle is \emph{falsified} if, at equal training data, telemetry-conditioned models show no gain on ego-motion and multi-view tasks.

\para{P2. Query-driven resolution (K1, K3, K11).}
Perception is the strongest facet for the median model on aerial benchmarks (HNS 0.55--0.60), while small and distant targets lose one to three quarters of their accuracy and have done so for a decade.
The token or pixel budget should follow the query and the ground sampling distance, rather than be spread uniformly across frames.
The principle is \emph{falsified} if, at equal token count, query-driven allocation does not raise recall of objects under 10 pixels over uniform compression.

\para{P3. Time as a continuum (K5, K6, K11).}
Aerial time spans frames, flights and satellite passes, yet each benchmark covers one scale, and long horizons are among the largest losses in the corpus (retention 0.34 across 41 trackers).
Benchmarks should include streaming and forward-looking questions, and span at least two time scales; models should keep state across them.
The principle is \emph{falsified} if single- or shuffled-frame baselines match full models on aerial temporal questions, which would also show that the questions are not temporal.

\para{P4. Platforms as a team (K8, K9).}
Multiple drones and air--ground pairs are the norm in operations, but appearance-based association fails across large viewpoint changes.
Views should be fused through a shared, geo-registered memory rather than concatenated as images.
The principle is \emph{falsified} if naive view concatenation matches geo-registered fusion on cross-view tasks.

\para{P5. Evaluate the construct (K4, K6, K7, K10).}
An aerial benchmark should demonstrate that its questions require the capability they name.
That requires per-task human and chance baselines, blind and single-frame floors, variance, evidence localization, and closed-loop transfer where the task implies action; \cref{sec:card} turns these into a card.
The principle is \emph{falsified} if benchmarks that report these controls rank models no differently from benchmarks that do not.

\section{The AVU-Card}
\label{sec:card}

\begin{table}[t]
\centering
\scriptsize
\setlength{\tabcolsep}{2.6pt}
\setlength{\aboverulesep}{0pt}\setlength{\belowrulesep}{0pt}
\renewcommand{\arraystretch}{1.12}
\rowcolors{2}{gray!20}{gray!2}
\renewcommand{\tabularxcolumn}[1]{m{#1}}
\begin{tabularx}{\linewidth}{>{\raggedright\arraybackslash}m{2.25cm} >{\raggedright\arraybackslash}X}
\toprule
\rowcolor{gray!50} \multicolumn{1}{c}{\textbf{AVU-Card field}} & \multicolumn{1}{c}{\textbf{AirCopBench~\cite{zha2026aircopbench}}} \\
\midrule
1 Facet-coded HNS & 14 tasks, 6 facets (13 tasks with human $-$ chance $\geq 20$); median-model perception 0.55, multi-view 0.50, ego-motion 0.46, spatial 0.45, planning 0.43, counting 0.11 \\
2 Chance and human & Per task (\cmark); human raters and subset size not stated in the main text; 1 task with human $-$ chance $=10.3$ \\
3 Blind, single-frame, shuffled floors & Blind models used only to filter questions; no floor reported \\
4 Variance & Not reported \\
5 Acquisition envelope & 2--6 drones; simulated and real views; altitude, GSD and view angle not reported per item \\
6 Inputs & Simultaneous still images; no telemetry; frame budget n/a \\
7 Time scales & None (single instant) \\
8 Evidence & Answer only; no localization of the supporting region or frame \\
9 Transfer & Fine-tuned and sim-to-real rows reported (+27.0, +19.6); no closed-loop test \\
10 Person identity & No identity or biometric task \\
\bottomrule
\end{tabularx}
\caption{\textbf{AVU-Card} fields with a worked example. Fields 1--4 make accuracy interpretable, 5--7 describe what makes the data aerial, 8--10 bound what the numbers can be used for.}
\label{tab:card}
\end{table}

P5 is operationalized as an one-table reporting card for aerial video understanding benchmarks (\cref{tab:card}).
It is deliberately short, since every field can be filled from quantities that a benchmark already computes or records, and none requires new annotation beyond a small human study.
Fields 1--4 would have made most analyses in this paper unnecessary, since they inform a reader whether an accuracy gain reflects the named capability.
Fields 5--7 expose the acquisition conditions of \cref{fig:retention}, which otherwise must be inferred from dataset descriptions, and state whether the model receives telemetry and at which time scales it is asked to reason.
Fields 8--10 bound the use of the numbers by recording whether answers are tied to evidence, whether competence transfers to closed-loop use, and whether the task involves identifying people.
The worked example shows that the best-documented peer-reviewed aerial MLLM benchmark leaves fields 3, 4 and 8 empty and fields 2 and 5 incomplete, which is a statement about the field's reporting norms rather than about that benchmark.

\para{Call to action.}
\textbf{Benchmark builders} should ship telemetry with every clip, fill the AVU-Card, and report results binned by altitude, ground sampling distance and object size.
\textbf{Model developers} should accept pose and ground sampling distance as inputs and report accuracy with and without them, so that P1 can be tested directly.
\textbf{Challenge organizers} should add telemetry-on/off and shuffled-frame tracks to established aerial challenges such as VisDrone.

\section{Alternative Views and Limitations}
\label{sec:discussion}

\para{``Scale and data will close the gap.''}
The strongest counter-position is that the gaps are transient.
The evidence does not rule out that larger or better-trained models will close them, but it bounds the cost, since at the observed slopes parameter scale alone would need several orders of magnitude (K2), and fine-tuning gains, although large, are uneven across facets, with the one peer-reviewed self-state task barely moving.
The position predicts that the next large gains will originate from geometric inputs and geometric supervision; a model that closes the ego-motion and cross-view gaps without either would falsify P1.

\para{``MLLMs are not meant for metric geometry.''}
A second view holds that geometry belongs to specialized modules, not to MLLMs.
The position is compatible with a modular design, but the benchmarks still need to test the composed system on geometric questions, and K4 shows that the peer-reviewed aerial MLLM benchmark does not.
Where an MLLM is given explicit 3D structure the gap largely closes, as on Ego3D-Bench an estimated cognitive map lifts InternVL3-8B from 43.1 to 56.0 (58.4 with true camera parameters) and a ground-truth map to 79.4, against 85.3 for humans~\cite{gholami2026ego3d}. This supports P1 rather than contradicting it.

\para{``The benchmarks are too recent to judge.''}
This is correct for the preprints, and the convergence rule treats them accordingly.
The claims that carry the position rest on peer-reviewed evidence, much of it from a decade of perception studies whose weaknesses have persisted (K11).

\para{Forensic search, long video and efficient detection.}
For forensic search, the low temporal IoU of correct answers (K7) means that retrieval should be scored on the evidence it returns, not only on the answer.
For long video, a retention of 0.34 from short- to long-term tracking (K5) indicates that state must be carried across the flight rather than recovered from sampled frames (P3).
For efficient UAV detection, resolution spent where the targets are pays off, since sliced inference without retraining raises small-object AP50 on VisDrone by 5.6--5.7 points for three detectors~\cite{Akyon2022SAHI}, which supports allocating the pixel budget by query and ground sampling distance (P2) rather than uniformly.

\para{Governance.}
Aerial video of people is subject to strict legal limits; in the European Union, real-time remote biometric identification in publicly accessible spaces for law enforcement is prohibited outside narrow exceptions (Art.~5 of the AI Act~\cite{euaiact2024}).
Forensic search over aerial footage should therefore default to non-biometric attributes such as clothing, vehicles, trajectories and events, and benchmarks should flag tasks that identify individuals (AVU-Card field 10).
In the corpus, person identity is the explicit target only of re-identification and person-search tasks~\cite{nguyen2025agvpreid,li2021uavhuman,park2026foresea}; none of the aerial MLLM benchmarks flags whether its questions involve identifiable people.

\para{Limitations.}
The analysis inherits the choices of the source papers, since human baselines are often measured on subsets (10--20\% of items) or by few raters, several contrasts rest on small subsets (the air--ground multi-view contrast covers a single task), and retention values for UAVDT tracking are extracted from bar charts.
Scale slopes use two or three sizes per family and are indicative, not a scaling law.
Only published numbers are used, so the proposed floors (blind, single-frame, shuffled) could not be measured on aerial benchmarks; this is the main test that the position leaves to future benchmarks.
Benchmark papers are written to expose weaknesses, so the corpus may over-represent hard conditions, although each retention contrast compares the same methods within one paper.
The evidence is not fully independent, since the VisDrone reports share organizers and data, while several models recur across benchmarks.
The model-matched comparison mixes prompts, frame sampling and answer formats that differ between benchmarks, so it supports the direction of K4 rather than the size of the differences; for 7 of the 8 cross-view pairs, the ground side is a single benchmark.

\section{Conclusion}
\label{sec:conclusion}

Examined together, a decade of aerial perception studies, the first peer-reviewed aerial MLLM and navigation benchmarks, ground-view analogues and recent preprints agree on a small set of claims.
Appearance is the strongest facet for typical models; accuracy is controlled by altitude, view, object size and time horizon; scale buys little; and competence does not survive the step from answering to acting.
The benchmarks cannot yet show whether their questions require video or geometry at all.
The gap is in the geometry, and so is the next step, which calls for models that are informed where the camera is and for benchmarks, documented with the AVU-Card, that reward using it.

\section*{Use of large language models}
Claude (Anthropic) was used to extract values from the benchmark papers, compute statistics, check grammar and generate the figures; all values are traceable to the cited sources, and the authors take full responsibility for all content.

{
    \small
    \bibliographystyle{ieeenat_fullname}
    \bibliography{bibliography}
}

\end{document}